\pdfoutput=1

\documentclass[11pt]{article}

\usepackage[preprint]{acl}

\usepackage{times}
\usepackage{latexsym}
\usepackage{color}
\usepackage{graphicx}
\usepackage{tabularx}
\usepackage{soul}
\usepackage{booktabs}
\usepackage{algorithm}
\usepackage{algpseudocode}
\usepackage{amsmath}
\usepackage{multirow} 

\usepackage{paralist}

\usepackage[T1]{fontenc}

\usepackage[utf8]{inputenc}

\usepackage{microtype}

\usepackage{inconsolata}

\title{ID-XCB: Data-independent Debiasing for Fair and Accurate Transformer-based Cyberbullying Detection}

\author{Peiling Yi \\
Queen Mary University of London \\
  \texttt{p.yi@qmul.ac.uk} \\\And
  Arkaitz Zubiaga \\
  Queen Mary University of London \\
  \texttt{a.zubiaga@qmul.ac.uk} \\}

\begin{document}

{\makeatletter\acl@finalcopytrue
  \maketitle
}
\begin{abstract}
Swear words are a common proxy to collect datasets with cyberbullying incidents. Our focus is on measuring and mitigating biases derived from spurious associations between swear words and incidents occurring as a result of such data collection strategies. 
After demonstrating and quantifying these biases, we introduce ID-XCB, the first data-independent debiasing technique that combines adversarial training, bias constraints and debias fine-tuning approach aimed at alleviating model attention to bias-inducing words without impacting overall model performance. We explore ID-XCB on two popular session-based cyberbullying datasets along with comprehensive ablation and generalisation studies. 
We show that ID-XCB learns robust cyberbullying detection capabilities while mitigating biases, outperforming state-of-the-art debiasing methods in both performance and bias mitigation. Our quantitative and qualitative analyses demonstrate its generalisability to unseen data.
\newline \textcolor{red}{Warning: This paper contains swear words, which do not reflect the views of the authors}.
\end{abstract}

\section{Introduction}

Cyberbullying is a form of bullying that takes place online \cite{Smith2008} and is defined as the repeated, deliberate aggressive behaviour by a group or individual towards a more vulnerable individual \cite{olweus2001bullying}.
Where cyberbullying is characterised by repeated aggression and power imbalance, 
research \cite{dadvar2012improved,dadvar2013improving,menin2021cyber,yi2022cyberbullying} suggests capturing these characteristics by modelling social media sessions \cite{yi2022session}, i.e. a series of conversational exchanges \cite{cheng2020session}, rather than from individual posts.

Cyberbullying detection models often suffer from biases leading to false positive predictions when a swear word is present \cite{agrawal2018deep,perera2021accurate,pamungkas2023investigating}, not least because swear words are often used as a proxy for data collection \cite{van2020multi}.
Swear words are however often used in other non-abusive contexts, not necessarily indicating cyberbullying incidents \cite{stephens2020swearing,lafreniere2022power}. A disproportionate presence of swear words across both positive and negative samples then leads to model overfitting, exhibiting a 
`swear word bias' \cite{hovy2021five}.

\begin{figure}[ht]
  \includegraphics[scale=0.5]{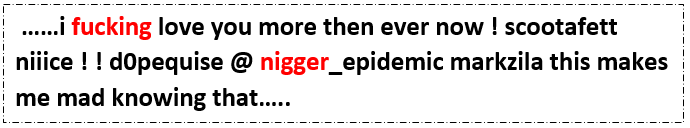}
  \caption{A snippet of false positive sample in Instagram }
  \label{fig:False}
\end{figure}

Figure \ref{fig:False} shows an example of a false positive bias due to swear word presence.

In the existing body of research debiasing text classification models, and particularly cyberbullying detection models, work has been limited to data-dependent constraints \cite{gencoglu2020cyberbullying,cheng2021mitigating}. These methods have proven satisfactory in mitigating biases for models tested on data from the same dataset or with similar characteristics, but risk overfitting on the seen data, limiting their generalisability and preserving both fairness and accurate performance.

\begin{figure}[tbh]
  \centering
  \includegraphics[width=\linewidth]{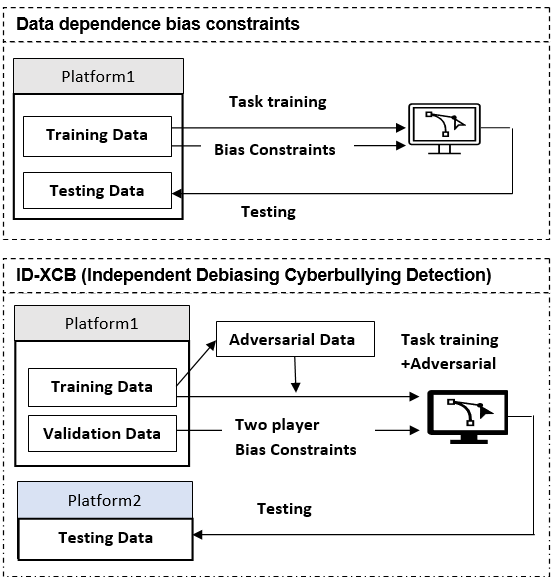}
  \caption{Data-dependent bias constraints vs. ID-XCB.}
  \label{fig:compare}
\end{figure}

Focusing on generalisable debiasing, we propose ID-XCB (Independent Debiasing  Cyberbullying Detection), the first data-independent debiasing method that avoids the need to see target data, in turn detaching the link between swear words and cyberbullying incidents (Figure \ref{fig:compare} illustrates data-dependent strategies vs ID-XCB). To achieve this, we integrate three strategies: (1) \textbf{adversarial training}, using adversarial examples and objective cost function to shift the model's focus away from profanity; (2) \textbf{Data In-dependence bias constraint optimisation} to work in the non-convex setting, as the downstream task is trained on training datasets but fairness constraints are derived from independent validation datasets; and (3) \textbf{contextualised embeddings} from transformer models, which support generalisability by transferring knowledge to the downstream task. Our contributions include:

\begin{compactitem}

\item We quantify swear word impact on transformer biases, showing its consistency across datasets and models.
\item We introduce ID-XCB, the first data-independent debiasing method for cyberbullying detection, which shows improved performance and bias reduction over state-of-the-art approaches on two datasets. It further shows potential for bias mitigation and performance trade-offs in challenging scenarios dealing with unseen data.

\item We perform in-depth analyses of specific swear words, model components, ablation experiments, and generalisation.
\end{compactitem}

\section{Related Work}

Cyberbullying detection is generally tackled as a binary classification task determining if an instance constitutes a case of cyberbullying or not. Recent research uses Transformer-based models to identify cyberbullying through multiple user interactions (i.e. sessions). \citet{gururangan2020don,yi2023session} demonstrate that transformer-based models can be strong, and competitive for session-based cyberbullying detection. However, researchers have shown that they can also suffer from degenerated and biased behaviour \cite{schramowski2022large}.



Existing work on debiasing text classification models can be categorised in four main directions: (1) statistically balancing training data, such as data augmentation \cite{dixon2018measuring}, sample weighting \cite{zhang2020hurtful}, identity term swapping \cite{badjatiya2019stereotypical} or injecting objective samples \cite{nozza2019unintended}; (2) mitigating embedding bias, such as fine-tuning pre-trained contextualised word embeddings \cite{kaneko2021debiasing} or using adversarial learning to reduce the bias \cite{sweeney2020reducing}; (3) proposing a multi-task learning model \cite{vaidya2019empirical} with an attention layer that jointly learns to predict the toxicity of a comment as well as the identities present in the comments in order to mitigate bias; and (4) inferring fairness constraints by using Error Rate Equality Difference to restrict the discrimination behaviour of the model \cite{zafar2017fairness}.

Model debiasing is understudied in cyberbullying detection, where studies to date have focused on feeding fairness constraints. For example, \citet{gencoglu2020cyberbullying} did so by using sentence-DistilBERT as a base model, adding the Fairness metric as a cost function to constrain bias during training. \citet{cheng2021mitigating} built on a reinforcement learning strategy relying on a pre-defined set of sensitive triggers to constrain a series of hierarchical attention networks. Current works focus on data-dependent fairness constraints, which satisfyingly reduce the false positive rate on the training data. However, if the terms of debiasing are strictly enforced on the training data, this may be beneficial to ensure fairness on similar data, but overfitting will also occur, thereby reducing the fairness of the model on unseen data \cite{hardt2016equality}.

\section{Datasets and Lexicon}

\textbf{Datasets} We use the two existing and widely-studied session-based cyberbullying datasets from two different social media platforms: Instagram \cite{hosseinmardi2015analyzing} and Vine \cite{rafiq2015careful}.
To collect social media sessions likely to contain cyberbullying events, authors of these datasets looked at the presence of toxic words to maximise the chances of collecting positive samples, which were subsequently manually annotated. Table \ref{tab:datasets} shows statistics of both datasets.

\begin{table}[ht]
  \centering
   \scalebox{0.8}{
  \begin{tabular}{lcc}
    \toprule
     & \textbf{Instagram(IG)}& \textbf{Vine(VN)}\\
    \midrule
    Cyberbullying Ratio & 0.29& 0.30 \\ 
    \# Sessions  & 2,218   & 970            \\
    \# Comments &159,277  &70,385           \\
    \# Users &72,176  &25,699           \\
     \# Avg. length of session              &900&698 \\
      \# Unique Swear words             &253 &207 \\
    \bottomrule
  \end{tabular}}
  \caption{Dataset statistics.}
  \label{tab:datasets}
\end{table}

\textbf{Lexicon} To determine the presence of swear words, we use the lexicon with 535 words provided in \cite{van2018automatic,badwords}.

\section{Swear Word Bias}

We adopt the Oxford Dictionary's definition of bias as the ``inclination or prejudice for or against one person or group, especially in a way considered to be unfair''. By `swear word bias', we refer to the impact of swear words during training on biasing model predictions, or model bias. A prominent swear word bias ultimately leads to the assignment of disproportionately high importance to the presence of swear words in model predictions.

\subsection{Distribution of swear words}

A first look at the distribution of swear words (Table \ref{tab:bias}) shows that cyberbullying events don't contain more swear words in cyberbullying detection datasets, likely limiting their utility for the predictions. (P(S|C)$\approx$P(S|NC)\&P(C|S)$\approx$P(NC|S)). In fact, around 70\% of the posts with swear words belong to the negative class, and fewer than 3\% of all sessions have no swear words in both datasets.

\begin{table}[h]
 \centering
    \scalebox{0.7}{
  \begin {tabular}{lcccccc}
    \toprule
     & P(C)&P(NC)&P(S|C)&P(S|NC)&P(C|S)&P(NC|S)\\
    \midrule
    Instagram& 0.29& 0.71&1.0&0.98&0.87&0.87 \\ 
    Vine& 0.30&0.69&1.0&0.97&0.86&0.84   \\
    \bottomrule
  \end{tabular}
   }
  \caption{Distribution of swear words (S).  C: Cyberbullying, NC: No Cyberbullying.}
  \label{tab:bias}

\end{table}

\subsection{Measuring swear word bias} \label{measure}

To measure a model's bias towards swear words, we adopt the Error Rate Equality Difference approach \cite{dixon2018measuring}, which relies on the FPR (false positive rate) and FNR (false negative rate) metrics, calculated as follows:

\begin{equation}
\text{FPR} =\frac{\text{FP}}{\text{FP}+\text{TN}} \hspace{1em}; \hspace{1em} \text{FNR} =\frac{\text{FN}}{\text{FN}+\text{TP}}
\end{equation} 

For each swear word $w$, this allows to compute the FPRD (FPR difference) and FNRD (FNR difference) as the model bias towards that word:

\begin{equation}
 \begin{aligned}
  \text{FPRD}_w=|\text{FPR}-\text{FPR}_w|\\
  \text{FNRD}_w=|\text{FNR}-\text{FNR}_w|
 \end{aligned}
\end{equation} 

Where $FPR$ and $FNR$ are calculated on the entire test set; $FPR_w$ and $FNR_w$ are calculated on the subset of the test set that contains $w$.

Having those, one can then aggregate the bias towards all swear words W under consideration, i.e. $FPED$ (false positive equality difference), and $FNED$ (false positive equality difference):

\begin{equation}
 \begin{aligned}
  \text{FPED} =\sum_{w\in W}|\text{FPR}-\text{FPR}_w|\\
  \text{FNED} =\sum_{w\in W}|\text{FNR}-\text{FNR}_w|
 \end{aligned}
\end{equation}  

Lower values indicate a fairer model.

\subsection{Quantifying bias with transformers}

Prior to moving on to bias mitigation, we first investigate and quantify the impact of swear words on model bias. Using a vanilla BERT model for initial experiments in both datasets, we show an analysis of the bias measurement of the most frequent swear words as well as most bias-inducing swear words in Table \ref{tab:popular}. This analysis shows that the most frequent swear words (left) have a low impact on model bias (left). If we look instead at the swear words that cause the highest model bias, we observe that these form a non-intersecting set of swear words that are less frequent (right). We will come back to these top bias-inducing swear words in our study to assess the effectiveness of debiasing strategies. In Appendix \ref{app:extended-bias} we show that this bias is consistent with five Transformer models.

\begin{table*}[!ht]
\centering
\begin{tabular}{lcc || lc | lc}
\hline
\multicolumn{3}{c||}{\textbf{Most frequent swear words}} & \multicolumn{4}{c}{\textbf{Top bias-inducing swear words}} \\
\hline
\textbf{SW} & \textbf{FPRD (IG)} & \textbf{FPRD (VN)} & \textbf{SW} & \textbf{FPRD (IG)} & \textbf{SW} & \textbf{FPRD (VN)} \\
\hline
fuck   & 0.008 & 0.007 & Piece of shit    & 0.929 & cunt    & 0.900 \\
shit   & 0.040 & 0.040 & nigger   & 0.929 & cum     & 0.500 \\
fuckin & 0.008 & 0.010 & dickhead & 0.929 & bitches & 0.400 \\
bitch  & 0.040 & 0.040 & gash     & 0.929 & boob    & 0.234 \\
hell   & 0.007 & 0.001 & cunt     & 0.429 & faggot  & 0.234 \\
\hline
\end{tabular}
\caption{Bias with most frequent swear words and with top bias-inducing swear words.}
\label{tab:popular}
\end{table*}

\section{ID-XCB}

In this section, we introduce ID-XCB, our debiasing method and its theoretical implementation in detail. The framework, depicted in Figure \ref{fig:architecture}, is divided into two parts:

\begin{figure}[tbh]
  \centering
  \includegraphics[scale=0.5]{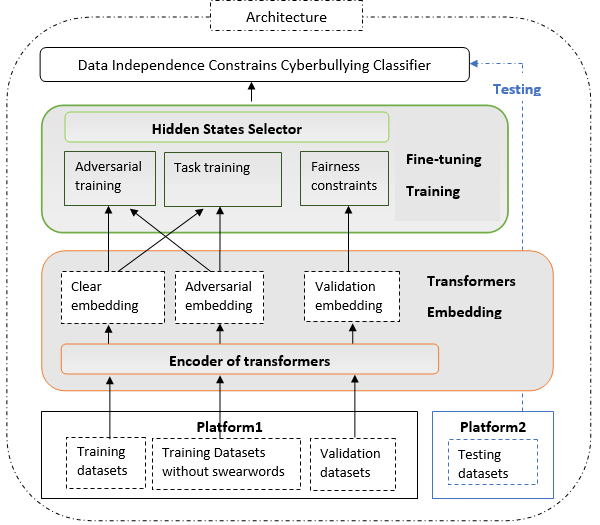}
 
  \caption{Architecture of ID-XCB.}
    \label{fig:architecture}
\end{figure}

\textbf{1. Transformers Embedding:} uses encoders of transformers to generate three kinds of embeddings for training: (i) \textbf{clear embeddings} from the original training dataset, (ii) \textbf{adversarial embeddings} from the training dataset but with all swear words replaced with a mask, and (iii) \textbf{validation embeddings} from validation datasets.

\textbf{2. Hidden states selector:} Compared to static embeddings, contextualised embeddings such as BERT, GPT and ELMo are less biased \cite{may2019measuring}, but still show a tendency to adopt biases during training \cite{zhao2019gender,kurita2019measuring}. These biases are learned in each layer \cite{bommasani2020interpreting}, thus fine-tuning the orthographic projections in intermediate (hidden) layers \cite{kaneko2021debiasing} is an efficient method that doesn't depend on the model. To identify which layer is good for fine-tuning, we add a hidden states selector, which iteratively extracts the representation of each layer as input to the classifier for fine-tuning, and then tests the generalisation and debiasing capabilities of each level.
 
\textbf{3. Fine-tuning Training:} is responsible for integrating three different training processes to break data dependency and improve the generalisability of the debiasing technique and the cyberbullying detection model, which includes adversarial training, tasking training and fairness training.   The hidden states selector helps find which layers are best to optimise knowledge.

The intuition of ID-XCB is that leveraging the combination of the three training loss functions will lead to improved debiasing and performance generalisation on unseen datasets. Details of the three loss functions are provided next.

\subsection{Training loss functions} \label{loss}

\subsubsection{Adversarial training}
\label{sssec:ilo}

The aim is to apply adversarial training against biased latent representations to mitigate unwanted bias. Thus, we utilise the cosine similarity function to generate cosine loss such that the model can't tell the difference between real training samples or artificially synthesised training samples. 
The loss function is defined as follows:

\begin{equation}
 \text{EmbeddingLoss}=1-\text{cos}(x_1,x_2) 
\end{equation}

Where $x_1$, and $x_2$ refer to the clear embeddings and adversarial embeddings. The training goal is to optimise EmbeddingLoss close to 0.

\subsubsection{Task training}

Binary Cross-Entropy loss is used to train the main task: cyberbullying detection. 

\begin{multline}
 \text{BCELoss}=y(\log\frac{(x_1+x_2)}{2})
\\+(1-y)(\log(1-\frac{(x_1+x_2}{2}))   
\end{multline}

Where $y$ refers to the training labels. The input is a synthetic embedding which is the average of clear and adversarial embeddings.

\subsubsection{Fairness constraints}
If the terms of debiasing are strictly enforced, it can benefit fairness but harm model accuracy. In practice, sensitive tuning is performed by using a proportional constraint, which can maintain a more suitable trade-off. We use an independent validation set to derive fairness constraints:


\begin{multline}
\text{FC}=\beta(\sum_{w=1}^n FPRD_w +\sum_{w=1}^n FNRD_w)
\label{al_9}  
\end{multline}

Where $\beta$ is how tightly the data is bound to adjust the fairness constraints, and $w$ is the swear word under consideration in the validation data.

\subsection{Constraint-based classifier}

It comprises two fully connected layers on top. The two-layer feed-forward network is designed with ReLU activation and 512 hidden sizes for the first layer and Softmax activation for the output layer. Batch normalisation is added to standardise these inputs and reduce the generalisation error, so as to increase the generalisability of the classifier.

\subsection{Joint training}


Following Algorithm \ref{alg:training}, we combine the three training losses in Section \ref{loss} to fine-tune a classifier.  Adversarial training and task training optimise ID-XCB model parameters on a training dataset, and simultaneously enforce the fairness constraints on a validation set to reduce swear word impact. However, as the FC loss is used on an independent validation set $V_v$ , this causes a non-convex combining loss. The non-zero-sum method \cite{cotter2019two} deals with non-differentiable, even discontinuous constraints.  
The training goal is not to converge the combined cost function to the lowest point, but to reach a certain trade-off. We set a threshold $t$ to achieve the training target.

\begin{algorithm}

	\caption{ID-XCB training. $V_s$: training embeddings; $V_a$: adversarial embeddings; $V_v$: validation embeddings; $Y_t$: training labels; $Y_v$: validation labels; $Y_a$: adversarial labels; $n$: number of epochs.}
	\begin{algorithmic}[1]
	\Require {$V_s$,$V_a$, $V_v$,$Y_t$,$Y_v$,$Y_a$,$n$}

		\For {epoch in range (n)}
                \For {$layer(i)=1,2,\ldots,12$}
			\For {step, ($V_si$,$V_ai$,$V_vi$),($Y_t$,$Y_a$)}
  	\State 
        $l_t$=BCELoss(($V_s$,$V_a$),$Y_t$) 
       \State $l_a$=EmbeddingLoss(($V_s$,$V_a$),$Y_a$)                    \State$ls_d$=FC(classifier,$V_v$)
    \If{($l_t$+$l_a$+$l_d$)$\geq t$}
       \State 
       classifier.backward($l_t$+$l_a$+$l_d$)               
        \State results[i]=classifier.evaluate($V_v$,$Y_v$)
        \Else
        \State 
        Exit!
	 \EndIf	 	
	 	\EndFor
     \EndFor
   \EndFor
  \end{algorithmic} 
 \label{alg:training}
\end{algorithm}



\section{Experiment Settings}


\textbf{Our models.} While ID-XCB is flexible and can adopt other transformer models, here we experiment with BERT\_base and RoBERTa\_base, which we refer to as ID-XCB$_{BERT}$ and ID-XCB$_{RoBERTa}$. When training these models, we use the training hyper-parameters recommended by \citet{Sun2019a}; Batch size: 16; Learning rate (Adam): $2e^{-5}$; Number of epochs: 4.

\textbf{Pre-processing.} We follow \citet{ge2021improving} to aggregate and clean session data, and to truncate session lengths to 512 tokens, with the difference that we do not perform oversampling (as our objective is to keep the original data imbalance).

\textbf{Baseline models.} We consider \citet{cheng2021mitigating} and \citet{gencoglu2020cyberbullying} for standing as highly influential debiasing methods for the cyberbullying detection task. Their approach of bias constraints on training data has consistently demonstrated state-of-the-art performance in recent experiments.
Thus, we compare our method with these two state-of-the-art cyberbullying detection debiasing methods (using BERT \cite{devlin2018bert} and RoBERTaA \cite{liu2019roberta} variants of those) as well as vanilla transformers:
(i) \textbf{De-RoBERTa \& De-BERT:} applied reinforcement on transformers \cite{cheng2021mitigating}; (ii) \textbf{FC-RoBERTa \& FC-BERT} which uses Error Rate Equality Difference to restrict transformers' discrimination behaviour \cite{gencoglu2020cyberbullying}; and (iii) \textbf{Roberta \& BERT.}



 
 

\begin{table*}[htb]
 \begin{center}
  \begin{tabular}{c|l||c|c|c|c|c}
  \toprule
  
    Source$\rightarrow$Target&\multicolumn{1}{c||}{Model}&F1&Rec.&Prec.&FPED&FNED\\
    \midrule
    \midrule
    \multirow{8}{*}{IG$\rightarrow$IG} & BERT &0.79&0.79&0.78&8.86&27.4\\
    & RoBERTa &0.86&0.87&0.86&6.81&23\\
    & De-BERT&0.84&0.82&0.83&14&20\\
    & De-RoBERTa &0.87&0.91&0.84&\textbf{3.2}&\textbf{13}\\
    & FC-BERT &0.75&0.92&0.81&5.05&16.4\\
    & FC-RoBERTa &0.81&0.83&0.80&15.2&16.1\\
    
    \cline{2-7}
    & ID-XCB$_{BERT}$ (ours) &0.86&0.87&0.85&6.7&16.3\\
    & ID-XCB$_{RoBERTa}$ (ours)  &\textbf{\underline{0.89}}&0.89&0.89&3.9&15.6\\
    \midrule
    \midrule
    \multirow{8}{*}{VN$\rightarrow$VN} & BERT &0.77&0.77&0.79&5.73&18\\
    & RoBERTa &0.85&0.82&0.91&4.66&16.99\\
    & De-BERT&0.74 & 0.71&0.86&10.4&\textbf{11.4}\\
    & De-RoBERTa&0.76 & 0.85&0.68&6.579&14.1\\
    & FC-BERT &0.66&0.67&0.66&16&17\\
    & FC-RoBERTa &0.79&0.81&0.76&8.1&13\\
    \cline{2-7}
    & ID-XCB$_{BERT}$ (ours) &0.83&0.83&0.85&3&17.1\\
    & ID-XCB$_{RoBERTa}$ (ours) &\textbf{\underline{0.89}}&0.86&0.93&\textbf{1.51}&11.73\\
   \bottomrule
  \end{tabular}
 \end{center}
 \caption{Results for ID-XCB and six baseline models.}
 \label{tab:id-results}
\end{table*}


\textbf{Ablated models.} Aiming to gain a better understanding of the contribution of each component of ID-XCB, we experiment with three ablated models: (i) \textbf{ID-XCB$_{EB}$} with no fairness constraints; (ii) \textbf{ID-XCB$_{BF}$} with no adversarial training; and (iii) \textbf{ID-XCB$_{EF}$} replacing the synthetic with the original embedding.


\textbf{Evaluation.}  We use five widely-used evaluation metrics for imbalanced datasets and model bias. These include (1) recall, precision and micro\-F1 as performance metrics, and (2) 
FPED and FNED, as fairness indicators, are cumulative deviation values for each swear word, the scale can be [0, Positive infinity], where 0 indicates no deviation.

\textbf{Train-test splits.} We choose 5 random folds with 80\%-20\% sessions for training / testing, reporting the average performances across the 5 runs.

\section{Results}

We next discuss results of our experiments and delve into numerous aspects of our model.

\subsection{Overall performance and debiasing}

Table \ref{tab:id-results} shows the results for all models, including our two ID-XCB variants and six baselines.
These results demonstrate the general superiority of our ID-XCB model, of which ID-XCB$_{RoBERTa}$ shows superior performance. We observe that in the in-dataset experiments it is capable for obtaining higher performance scores. It also achieves superior bias mitigation than most of the models, showing the best FPED score across all models for the Vine dataset. It is also only behind De-BERT for FNED in the Vine dataset and only behind De-RoBERTa in the Instagram dataset. 

Interestingly, we observe that ID-XCB achieves competitive debiasing while leading to improved overall performance. Debiasing can often sacrifice performance, especially when applying data rebalancing, regularisation, and adversarial learning debiasing methods \cite{chen2023bias}. In this challenging scenario, ID-XCB achieves a trade-off between debias and performance.



\begin{table}[htb]
 \begin{center}
  \begin{tabular}{c|l||c|c|c}
  \toprule
  
    Src$\rightarrow$Tgt&\multicolumn{1}{c||}{Model}&F1&FPED&FNED\\
     \midrule
   \multirow{4}{*}{IG$\rightarrow$IG}& ID-XCB & \textbf{0.89} & \textbf{3.9} & \textbf{15.6} \\
    \cline{2-5}
    & ID-XCB$_{EB}$ &0.85&23&16\\
    & ID-XCB$_{BF}$ &0.86&10&21\\
    & ID-XCB$_{EF}$ &0.84&15&20\\
    \midrule
     \midrule
   \multirow{4}{*}{VN$\rightarrow$VN}& ID-XCB & \textbf{0.89} & \textbf{1.51} & \textbf{11.73} \\
    \cline{2-5}
    & ID-XCB$_{EB}$ &0.87&6.46&16.38\\
    & ID-XCB$_{BF}$ &0.88&4.98&16.88\\
    & ID-XCB$_{EF}$ &0.86&6.71&16.9\\
   
   \bottomrule
  \end{tabular}
 \end{center}
 \caption{ID-XCB$_{RoBERTa}$ vs ablated variants.}
 \label{tab:ablation}
\end{table}

\subsection{Ablation study}

We conduct ablation tests without adversarial training, fairness constraints and synthetic embedding in ID-XCB$_{RoBERTa}$. The results in Table \ref{tab:ablation} show that all components contribute to a noticeable improvement to both performance and debiasing. When the synthetic embedding (ID-XCB$_{EF}$) is replaced by the original embedding, the bias increases sharply. This aligns with our expectations, as the component was designed to mitigate the model's fixation on specific swear words, but still preserving their contribution to the task.

\subsection{Revisiting bias-inducing swear words}

Table \ref{tab:popular} shows the top 5 most bias-inducing swear words for both datasets when using a vanilla BERT model, which we revisit here to assess the extent to which ID-XCB helps mitigate the bias induced by these words. Table \ref{tab:de-popular} shows the FPRD scores for these top 5 swear words in Instagram and Vine, respectively. We observe a significant decrease in bias score (FPRD) in both cases for all the words.

\begin{table}[h]
  \centering
  \small
\begin{tabular}{llcc }
\toprule
 & \textbf{SW} & \textbf{RoBERTa} & \textbf{ID-XCB} \\
\midrule
\multirow{5}{*}{IG} & piece of shit  & 0.929 & 0.449 \\
 & nigger   & 0.929 &0.135 \\
 & dickhead & 0.929& 0.706\\
 & gash     & 0.929 &0.049  \\
 & cunt     & 0.429 & 0.363\\
\midrule
& Average & 0.829 & 0.340 \\
\midrule
\multirow{5}{*}{VN} & cunt    & 0.900&0.242 \\
 & cum     & 0.500 &0.026\\
 & bitches & 0.400 &0.338\\
 & boob    & 0.234&0.075 \\
 & faggot  & 0.234&0.214 \\
\midrule
& Average & 0.454 & 0.179 \\
\bottomrule
\end{tabular}
\caption{Reduced bias (FPRD) with ID-XCB$_{RoBERTa}$.}
\label{tab:de-popular}
\end{table}



\subsection{Generalisation}

\begin{table*}[htb]
 \begin{center}
  \begin{tabular}{c|l||c|c|c|c|c}
  \toprule
  
    Source$\rightarrow$Target&\multicolumn{1}{c||}{Model}&F1&Rec.&Prec.&FPED&FNED\\
    \midrule
    \midrule
    \multirow{8}{*}{IG$\rightarrow$VN} & BERT &0.54&0.52&0.61&4&18\\
    & RoBERTa &0.70&0.71&0.701&28.3&31.9\\
    & De-BERT &0.64&0.68&0.63&14&30\\
    & De-RoBERTa &0.62&0.71&0.60&32&27\\
    & FC-BERT &0.41&0.50&0.34&\textbf{3}&0\\
    & FC-RoBERTa &0.57&0.65&0.54&7.2&\textbf{10.9}\\
    \cline{2-7}
    & ID-XCB$_{BERT}$ (ours) &0.73&0.72&0.74&23&31\\
    & ID-XCB$_{RoBERTa}$ (ours) &\textbf{\underline{0.76}}&0.78&0.75&20.3&17.4\\
    \midrule
   \midrule
   \multirow{8}{*}{VN$\rightarrow$IG} & BERT &0.70&0.66&0.79&24&37\\
   & RoBERTa &0.71&0.69&0.79&32.4&35.4\\
   & De-BERT&0.64&0.68&0.63&31&30\\
   & De-RoBERTa &0.61&0.59&0.69&12&\textbf{15}\\
   & FC-BERT &0.31&0.53&0.63&13.6&3.5\\
   & FC-RoBERTa &0.41&0.50&0.34&\textbf{5}&16\\
   \cline{2-7}
   & ID-XCB$_{BERT}$ (ours) &0.74&0.73&0.75&20&34\\
   & ID-XCB$_{RoBERTa}$ (ours) &\textbf{\underline{0.81}}&0.80&0.81&23&35\\
   \bottomrule
  \end{tabular}
 \end{center}
 \caption{Cross-dataset results for ID-XCB and six baseline models.}
 \label{tab:cd-results}
\end{table*}

\begin{figure}[h]
  \centering
  \includegraphics[width=\linewidth]{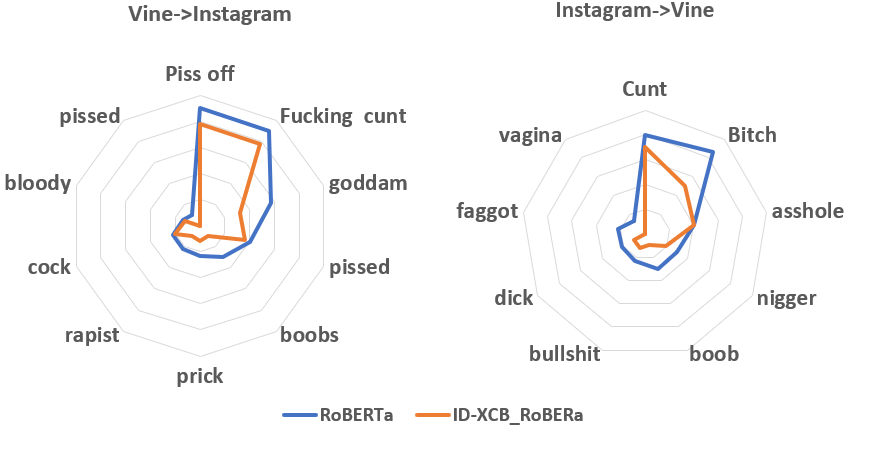}
  \caption{Swear word bias in RoBERTa vs ID-XCB$_{RoBERTa}$ for cross-platform experiments.}
  \label{fig:swearbias}
\end{figure}

We assess the generalisability of ID-XCB by testing on unseen data. Looking again at the highest bias-inducing words, Figure \ref{fig:swearbias} demonstrates that ID-XCB smoothly shrinks the highest bias at the word level without stimulating new bias, proving its effectiveness on mitigating bias for unseen data.

Table \ref{tab:cd-results} shows results for cross-platform experiments.
Despite performance improvement with ID-XCB$_{RoBERTa}$, improvement on debiasing is inconsistent, which we explore next by discussing the trade-off between the two metrics.

\subsection{Assessing the trade-off between performance and debiasing}

The trade-off between performance and debiasing is important as we observe that, for example, lowest performing models, FC-RoBERTa (19\% and 40\% below our model in the two datasets), achieve some of the best debiasing scores. Given the lack of a joint metric, we analyse the interplay between them. Therefore, we design constraint weights and layer selectors to achieve a balance between debiasing and maintaining performance.
Figure \ref{fig:swearbias} depicts the (FPRD) value for each highly biased word in the cross-platform task with four models. We use radar plots for multivariate comparisons to visualize that
our constraints don't promote new biases generated during cross-platform tasks. Models smoothly shrink the highest bias in word level without stimulating new biases.

In Table \ref{tab:cd-results}, we found that the model with the worst cross-platform performance (F1) also had the lowest bias value. Because FPED and FNED rely on differences in deviations and appear to differ minimally if they are both very large across the dataset and bias triggers. For example, FC-BERT model on VN$\rightarrow$IG and IG$\rightarrow$VN. As pointed out by \citet{borkan2019nuanced}, these are metrics of fairness rather than measurements of performance. It is difficult to find a simple linear relationship between task performance and debiasing effects.

To improve performance, our component increases the debiasing of the training set. However, the data distribution of the training set and the test set are completely different on cross-platform tasks. Overly heavy debiasing will improve the cross-platform performance of the system, but it will also increase the bias in unseen datasets. How to strike a balance is discussed next.

\subsubsection{Impact of constraint weighting}

\begin{figure}[htb]
  \centering
  \includegraphics[width=\linewidth]{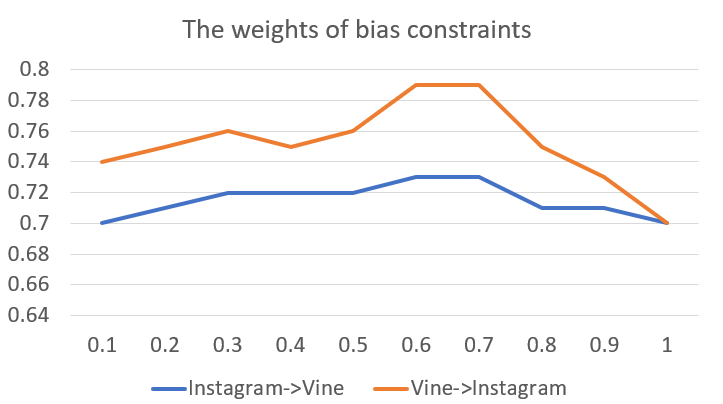}
  \caption{Impact of constraint weighting. X axis for constraint weights ($\beta$) and Y axis for F1 score.}
  \label{fig:weights}
\end{figure}

The weight determines how many constraints are applied to the bias.
We assess 10 values ranging from [0.1-1] at intervals of 0.1. 
Figure \ref{fig:weights} illustrates the correlation between bias constraint weights and task performance. Both cross-dataset settings show the same pattern: gradually debiasing on the training dataset enhances the model's performance on unseen datasets. The model achieves optimal performance when parameters are set to 0.6 or 0.7. However, tightening the constraints beyond this point leads to a significant drop in model performance. This demonstrates that excessive debiasing during training may not generalise well.

\subsubsection{The bias through different layers}

\begin{figure}[htb]
  \centering
  \includegraphics[width=\linewidth]{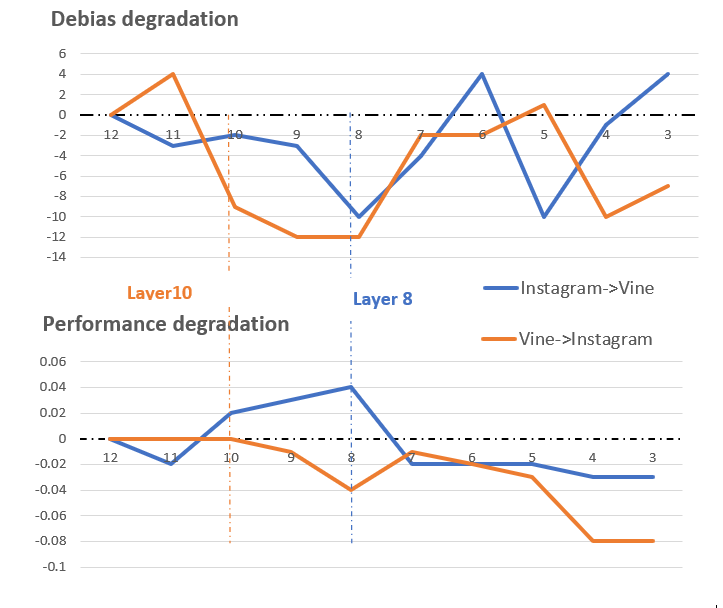}
  \caption{Bias across layers: X axis refers to 12 transformer layers. Y axis refers to relative performance and debiasing compared to layer 12.}
  \label{fig:layer}
\end{figure}

The hidden state selector is used to select the most transferable layer in transformers, so as to improve the transferability of the overall model.
To gain a deeper understanding into how the performance of various layers impacts debiasing and cyberbullying detection transferability, we measure the F1 score and FPED in each layer, fixing the last layer as the reference. Figure \ref{fig:layer} shows that layer 10 on IG$\rightarrow$VN leads to the best performance-debiasing trade-off, and layer 8 on VN$\rightarrow$IG.  We conclude two interesting phenomena: 1) A non-linear relationship exists between debiasing and performance. For instance, the eighth and fifth layers demonstrate the smallest FPRD values (IG$\rightarrow$VN), yet their performance reaches two extremes; 2) The first few layers focus on learning general features, rendering debiasing less impactful on overall model performance. However, in deeper layers, the network pays more attention to specific features, making debiasing efforts more effective in enhancing performance.

\section{Conclusion}

We introduce ID-XCB, the first data-independent debiasing approach for cyberbullying detection.
After quantifying biases with existing methods, we show the effectiveness of ID-XCB on two datasets, achieving competitive bias constraints enabling data-independent debiasing and training. This improvement is generally consistent in in-dataset experiments, whereas the improvement is particularly on performance for cross-dataset settings, with a good balance on debiasing. Our work in turn establishes a new benchmark addressing the lack of previous research on bias mitigation in cross-platform cyberbullying detection.

\section{Limitations}

Our work is however not without limitations. Most importantly, the dearth of available datasets leads to inevitable limitations in further studying generalisability across a more diverse set of datasets and across other social media platforms beyond Instagram and Vine. While there has been a more substantial body of work in related tasks within the umbrella of online abuse detection, such as hate speech detection, research on cyberbullying detection, and particularly on session-based cyberbullying detection, is much more limited to date and would greatly benefit from access to a broader collection of available datasets.

Our proposed ID-XCB model demonstrates state-of-the-art performance on the cyberbullying detection task, enabling some generalization across different datasets and platforms. This improvement comes with a competitive trade-off on performance and debiasing, however the model is not always consistently best across both metrics, which shows an area for further improvement. 

\section{Ethics Statement}

The aim of our research is to contribute to society and to human well-being by curbing incidents of cyberbullying online and particularly on social media. Our approach to mitigating biases  in keyword presence across cyberbullying and non-cyberbullying incidents is performed without leveraging any identity information to support the debiasing, and hence our approach is designed to support all individuals equally and with no intended discrimination or potential for harm towards any vulnerable groups.

There is an inevitable risk, as adversaries, who actually engage in cyberbullying incidents, may use this kind of research for malicious purposes such as to learn how to circumvent detection. This is however not the intended use of our research.



	
		 	




\bibliography{cyberbullying_bias}

\appendix

\section{Extended quantification of biases with five transformer models}
\label{app:extended-bias}

\begin{figure*}[h]
  \centering
  \includegraphics[scale=0.8]{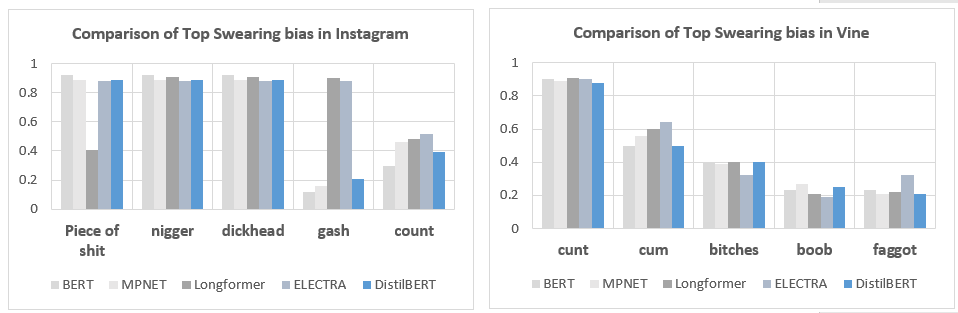}
  \caption{Swear word bias for 5 transformer models on both datasets.}
  \label{fig:Biase}
\end{figure*}

Experiments involving five transformer models (BERT,
MPNET,
Longformer,
ELECTRA,
DistilBERT)
reveal different levels of bias, consistently displaying higher bias scores across both datasets (Figure \ref{fig:Biase}).

\end{document}